\documentclass[letterpaper, 10 pt, conference]{ieeeconf}  

\IEEEoverridecommandlockouts                              

\def\shrug{\texttt{\raisebox{0.75em}{\char`\_}\char`\\\char`\_\kern-0.5ex(\kern-0.25ex\raisebox{0.25ex}{\rotatebox{45}{\raisebox{-.75ex}"\kern-1.5ex\rotatebox{-90})}}\kern-0.5ex)\kern-0.5ex\char`\_/\raisebox{0.75em}{\char`\_}}}
\usepackage{graphics} 
\usepackage{epsfig} 
\usepackage{mathptmx} 
\usepackage{times} 
\usepackage{amsmath} 
\usepackage{amssymb}  
\usepackage{color}
\usepackage{float}
\usepackage{cite}
\usepackage{svg}
\usepackage{titlesec}
\usepackage{graphicx}
\usepackage{subcaption}
\usepackage{caption}
\usepackage{pdflscape}
\usepackage{textcomp}\usepackage{gensymb}
\usepackage{booktabs}
\usepackage{array}
\usepackage{multirow}
\usepackage{colortbl}
\usepackage[dvipsnames]{xcolor}
\definecolor{black}{rgb}{0,0,0}
\definecolor{white}{rgb}{1,1,1}
\definecolor{darkred}{rgb}{0.5,0,0}
\definecolor{darkgreen}{rgb}{0,0.5,0}
\definecolor{darkblue}{rgb}{0,0,0.5}
\usepackage{nicefrac}
\usepackage{upgreek}
\usepackage{isomath}
\usepackage{fancyhdr}

\usepackage{units}
\usepackage{rotating}
\usepackage{url}
\usepackage{dblfloatfix}
\usepackage{makecell}
\title{\LARGE \bf
Learning Multi-Agent Local Collision-Avoidance for Collaborative Carrying tasks with Coupled Quadrupedal Robots 
}

\author{Francesca Bray$^{1}$, Simone Tolomei$^{2}$, Andrei Cramariuc$^1$, Cesar Cadena$^{1}$, Marco Hutter$^{1}$
\thanks{$^{1}$Robotic Systems Lab, ETH Zurich. Email: {\tt\small frbray@ethz.ch}.  $^{2}$ Centro di Ricerca “Enrico Piaggio”, and Dipartimento di Ingegneria dell’Informazione, Università di Pisa, Pisa, Italy.}
\thanks{This work was supported by the Luxembourg National Research Fund (Ref. 18990533), the European Union’s Horizon Europe Framework Programme under grant agreement No 101121321, the Swiss National Science Foundation  (SNSF) through the National Centre of Competence in Digital Fabrication (NCCR dfab) and project No. 200021E\_229503.
This work has been conducted as part of ANYmal Research, a community to advance legged robotics. }%
\thanks{We thank Dylan Vogel and William Talbot for their helpful insights and support.}%
}

\begin{document}

\maketitle
\pagestyle{empty}

\begin{abstract}
Robotic collaborative carrying could greatly benefit human activities like warehouse and construction site management. However, coordinating the simultaneous motion of multiple robots represents a significant challenge. Existing works primarily focus on obstacle-free environments, making them unsuitable for most real-world applications. Works that account for obstacles, either overfit to a specific terrain configuration or rely on pre-recorded maps combined with path planners to compute collision-free trajectories. This work focuses on two quadrupedal robots mechanically connected to a carried object. We propose a Reinforcement Learning (RL)-based policy that enables tracking a commanded velocity direction while avoiding collisions with nearby obstacles using only onboard sensing, eliminating the need for precomputed trajectories and complete map knowledge. Our work presents a hierarchical architecture, where a perceptive \textit{high-level object-centric} policy commands two pretrained locomotion policies. Additionally, we employ a game-inspired curriculum to increase the complexity of obstacles in the terrain progressively. We validate our approach on two quadrupedal robots connected to a bar via spherical joints, benchmarking it against optimization-based and decentralized RL baselines. Our hardware experiments demonstrate the ability of our system to locomote in unknown environments without the need for a map or a path planner. The video of our work is available in the multimedia material.
\end{abstract}

\vspace{-10pt}
\section{Introduction}
\label{intro}
Collaborative object transportation is a fundamental aspect of human interaction, particularly in scenarios where a single individual cannot manipulate an object due to its size or weight. 
Such tasks are common in practical domains, including construction sites and warehouse operations, where coordinated motion and spatial awareness are required to avoid collisions with environmental obstacles. 

Examples of robotic collaboration are becoming more and more frequent in recent years, proposing both \textit{model-based} and \textit{model-free} approaches.
\textit{Model-based} ones \cite{tracked_robots,mini_wheeled, ccma, hirche1, hirche2,multi_arms,quadruped_with_arms,quadruped_cent_dis_approaches, Vincenti-RSS-23} require a mathematical model of the whole system that describes not only the single agents' dynamics, but also that of the whole team.
\begin{figure}[ht!]
    \centering
    \includegraphics[width=\columnwidth]{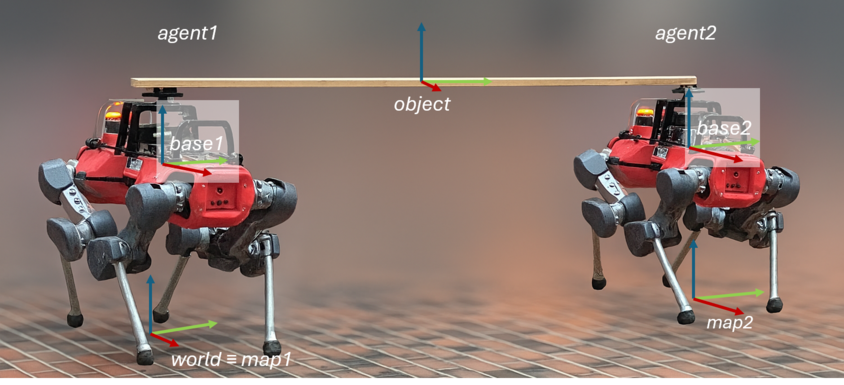}
    \caption{System deployed for hardware experiments.
    Each quadrupedal robot is provided with a mount ending with a spherical joint  connected to a bar of 2 m length.
    The task-relevant frames are highlighted.}
    \vspace{-20pt}
    \label{fig:system}
\end{figure}
The complexity of these models increases with the number of agents involved and the individual robots' characteristics, making them impractical for online optimization solvers.
As a result, simplified models are frequently adopted \cite{Vincenti-RSS-23, quadruped_cent_dis_approaches}, which may reduce robustness to unmodeled or unforeseen external influences.
On the other hand, \textit{model-free} approaches \cite{biped,anymal_b_raisim, quadruped_tethered, push_hierarchical} do not rely on accurate system models, and the use of domain randomization during training can improve their robustness to external disturbances.
Moreover, legged robots have successfully been shown to push a large object to a desired goal location \cite{push_hierarchical} using a hierarchical RL policy architecture.
Effective collaborative carrying of extended objects with quadrupedal and bipedal robots has been shown in \cite{biped,anymal_b_raisim}, while \cite{quadruped_tethered} presents a team of two legged robots carrying a heavy weight tethered to the agents' bases.

However, all the above-mentioned works execute collaborative manipulation tasks in empty environments, which is often not true for complex real-world scenarios. 
To approach this limitation, we employ an \textit{object-centric} hierarchical RL controller able to track a commanded velocity direction while avoiding collisions with the system's surroundings.
In particular, the contribution of this work consists of a local collision-avoidance policy for a system of mechanically coupled quadrupedal robots that operates in unknown environments without requiring prior map knowledge or external path planners.
To validate our approach, we benchmark it against optimization-based and decentralized RL baselines.
Finally, we demonstrate the efficacy of our approach through hardware experiments across a variety of static obstacle configurations, as well as in a representative dynamic scenario.

\section{Related work}
\label{sec:related_work}
In the following, we review existing approaches that address obstacle presence in the environment.

\subsection{\textit{Model-based}}
\label{sec:mpc}
The work presented in \cite{control_barrier_function} demonstrates a system composed of two quadrupedal robots mechanically connected to a bar via ball joints. 
The authors propose a control architecture with three nested layers, where the highest layer consists of a planner with full access to obstacle locations and dimensions. 
This planner computes safe trajectories for the robots' bases while accounting for the system’s kinematic constraints. 
The intermediate layer is a \textit{model-based} controller that represents the system as a team of interconnected Single Rigid Bodies (SRBs) and optimizes its dynamic motion to track the planned trajectories. 
The resulting optimal ground reaction forces are then passed to the lowest layer, a Model Predictive Control (MPC) locomotion controller, which tracks them at the joint level.

Similarly, \cite{florian_jrc} employs a planner to generate collision-free trajectories for a team of holonomic wheeled platforms equipped with manipulators.
With full access to obstacle positions and dimensions, the planner computes desired paths for both the robot bases and their end-effectors.

In \cite{javier}, the authors present a two-tier framework for multi-robot formation control in dynamic environments. 
Utilizing full map knowledge, the global planner identifies a long-range path by sampling large, obstacle-free convex regions and connecting them via feasible formation configurations. 
To handle real-time execution, a local planner operates in position-time space, growing local convex polytopes to account for moving obstacles while optimizing formation parameters to track the intermediate configurations provided by the global path. 
This local optimization abstracts robot dynamics, which are subsequently handled at a higher frequency by individual low-level controllers that ensure inter-robot collision avoidance and adherence to the optimized formation.

Generally, \textit{model-based} controllers require explicit modeling of the full system and its components, which often results in overly complex or inaccurate representations. 
Hence, we favor a \textit{model-free} approach.

\vspace{-5pt}
\subsection{\textit{Model-free}}
\label{sec:rl}
The work presented in \cite{push_obstacle_RRT_planner} demonstrates a team of quadrupedal robots collaboratively pushing an object from an initial to a target position.
At the highest level, a Rapidly-exploring Random Tree (RRT)-based planner generates a collision-free trajectory for the object, leveraging full knowledge of the environment and obstacle locations. 
Adaptive policies extract subgoals along this path and forward them to a mid-level pushing controller, which computes the desired object-centric velocity commands for the agents. 
These commands are then executed by a low-level locomotion policy.

In \cite{obstacles_2d_DRL}, two holonomic wheeled robots are trained to transport a long beam through a narrow aperture. 
However, this work focuses on a single map configuration only, hence overfitting to it.

Finally, \cite{obstacles_cnn} also deploys a team of wheeled robots carrying a beam-like payload to be transported through a door. 
Their system selects among three discrete high-level motion types using a Convolutional Neural Network (CNN)-based Q-learning framework.
The need to manually define these action modes can be limiting, as it requires the explicit specification of coordination strategies beforehand. \\

Generally, the reliance on a prerecorded map limits the applicability of such controllers in real-world scenarios, where terrain configurations may frequently change and dynamic obstacles are often present. 
Hence, in this work we present a collaborative carrying system able to track a commanded velocity direction in unknown environments while locally avoiding collisions with its surroundings.

\begin{figure*}
    \centering
    \includegraphics[width=0.8\textwidth]{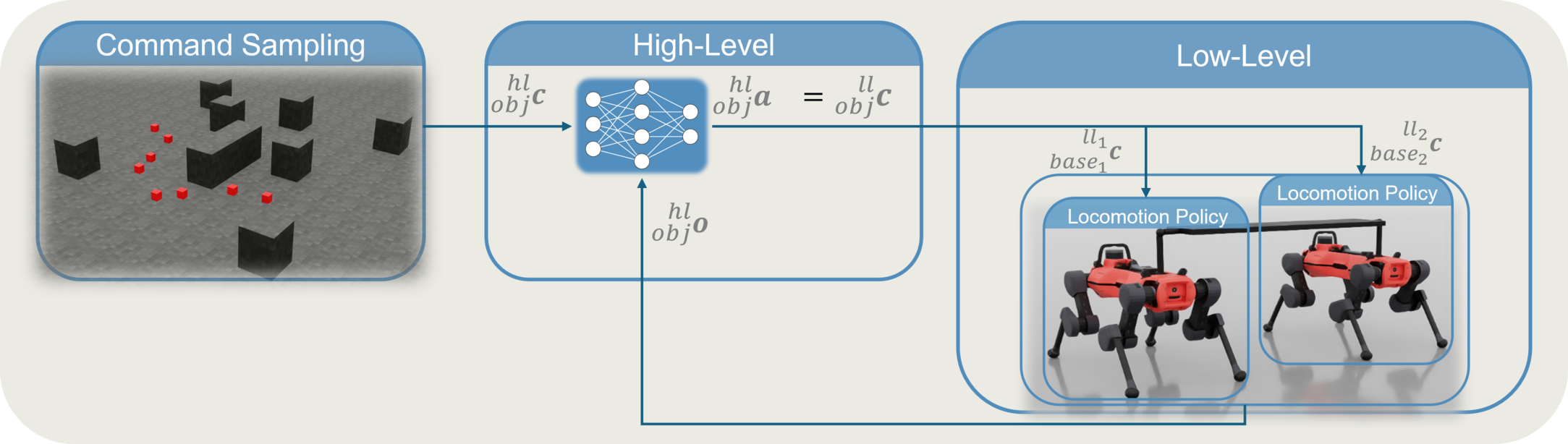}
    \caption{Policy architecture. 
    Single-robot feasible paths are sampled from the terrain, and the direction of the next waypoint defines the \textit{high-level} command  $^{hl}_{obj}\mathbf{c}$. 
    Together with system-level observations, $^{hl}_{obj}\mathbf{o}$, these are provided to the \textit{high-level} policy, which outputs \textit{high-level} actions $^{hl}_{obj}\mathbf{a}$.
    These represent the rotated \textit{low-level} commands $^{ll_i}_{base_i}\mathbf{c}$, which are executed by the trained locomotion policies.}
    \vspace{-15pt}
    \label{fig:architecture}
\end{figure*}

\section{Method}
\label{sec:method}

\subsection{Problem Statement}
\label{sec:prob_statement}
To study the problem of collaborative robotic transportation we deploy the system shown in Fig.\ref{fig:system}, where two quadrupeds robots are connected to a bar of length $L$ via a ball joint, which allows independent yaw rotation beneath the object, as well as limited roll and pitch motion relative to it.

We formulate the problem as finding a controller $\pi$ that at every time step \textit{t} maps proprioceptive $\textbf{o}_{sys}^t$ and exteroceptive $\textbf{z}^t$ observations to joint-level actions $\textbf{a}^t$ that produce coordinated motion of the entire system:  $\pi(\textbf{o}_{sys}^t, \textbf{z}^t) = \textbf{a}^t$.
The controller observes the object and agents' base velocities ${\boldsymbol{\nu}}_{obj}^t$, ${\boldsymbol{\nu}}_{base_i}^t,$ $i = 1,2$, along with the relative orientation between each agent and the bar $\psi_{obj, base_i},$ $i = 1,2$.
These proprioceptive inputs are collected by each agent via its onboard default state estimator, and provide information about the current state of the system.
These are crucial for coordinating agents motion and minimizing internal forces on the object.
In addition, each robot is equipped with four depth cameras, which are used to construct an elevation map of the system's surroundings with dimensions $H \times W$ and resolution $h_{res}$, which is then flattened into the vector $\textbf{z}^t \in \mathbb{R}^{H \times W}$.
These exteroceptive observations enable the system to perceive obstacles and avoid collisions by estimating the minimum distances between the environment and the system $d_{min_{obj}}$, $d_{min_i},$ $i = 1,2$.

To address the formulated problem, we employ a hierarchical RL structure consisting of\textit{ high-} and \textit{low-level} controllers, as illustrated in Fig. \ref{fig:architecture}.
In the following, we describe the components of the \textit{high-level} policy.

Throughout this work, we use the notation $^l_bx_a$ to denote the variable $x$ of frame \textit{a} expressed in frame \textit{b}. Moreover, the left superscript $^l$ is used to distinguish between \textit{high-} and \textit{low-level} variables.
Finally, the values of all parameters introduced in this section are summarized in Table \ref{tab:appendix_param} in the Appendix.

\subsection{Terrain and Curriculum}
\label{sec:terrain}
 \begin{figure}[ht!]
    \centering
    \includegraphics[width=\columnwidth]{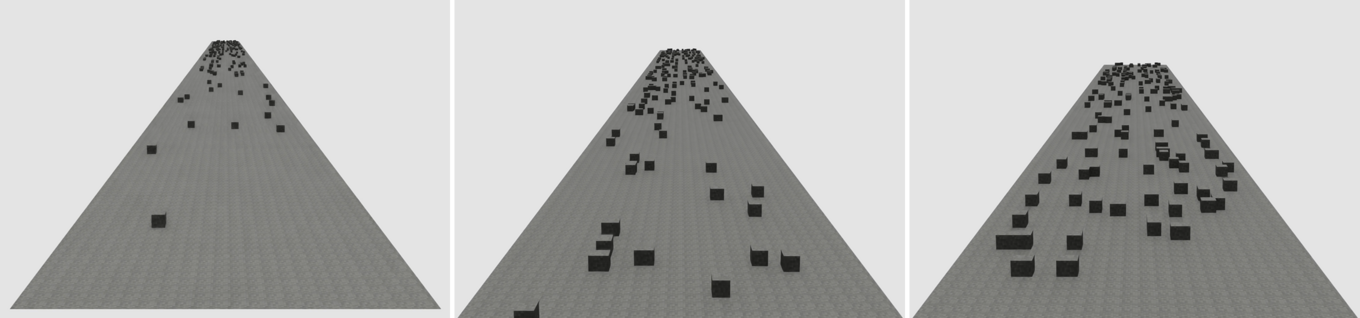}
    \caption{Example of a generated terrain used for training. 
    The leftmost picture corresponds to the easiest curriculum levels, while the rightmost represents the more challenging ones.}
    \vspace{-15pt}
    \label{fig:terrain}
\end{figure}
To progressively increase task difficulty, we employ a game-inspired curriculum approach, which has been demonstrated to enhance the learning process \cite{nikita_minutes}.
At the beginning of each training, we generate a terrain comprising $N_C$ levels of progressively increasing difficulty, ranging from 0 to $D_{max}$.
For each resulting subterrain, the difficulty level $D_i$ corresponds to the percentage of occupied terrain, with randomly placed obstacles.
These are generated as boxes of height $H_{obs}$, while width and length are in the range: [$s_{min}$, $s_{max}$] m.
Fig. \ref{fig:terrain} illustrates an example of the resulting terrain, with the easiest curriculum levels on the left and the most challenging ones on the right.
Agents can move up or down the curriculum depending on their performances (see Sec. \ref{sec:commands}).
\vspace{-5pt}
\subsection{Commands}
\label{sec:commands}
At the beginning of each training episode, we sample $n_p$ points over free space, ensuring that each point maintains a minimum safety distance $d_{s,wp}$ from the nearest obstacle.
This distance is chosen based solely on the dimensions of a single quadruped and is independent of the bar length $L$.
Once the sampling process is over, we apply \textit{Dijkstra's} algorithm to compute all available shortest paths, retaining only $N_P$ of them with lengths in the range [$l_{min}, l_{max}$].

We define the \textit{high-level} command $^{hl}_{obj}\textbf{c}$ as the desired velocity direction of the \textit{object} frame in the \textit{xy}-plane.
We do not specify a desired angular velocity, as we want it to be learned by the \textit{high-level} policy.
The command $^{hl}_{obj}\mathbf{c}$ is then defined as the norm of the distance in the \textit{xy}-plane between the next waypoint $_W\mathbf{r}_{wp_{xy}}$ and the current position of the object frame $_W\mathbf{r}_{obj_{xy}}$:
\begin{equation}
   ^{hl}_{obj}\mathbf{c} = \xi_{obj,W}||(_W\mathbf{r}_{wp_{xy}} - _W\mathbf{r}_{obj_{xy}})||,
\end{equation}
where $\xi_{obj,W}$ represents the rotation quaternion between the frames \textit{world} and \textit{object}.
Once the \textit{object} frame is within a radius $R_{wp}$ from one waypoint, that node is considered reached, and the directional command starts pointing towards the next available one.
If the goal is reached, the episode is terminated and the environment reset.

Each sampled path is assigned the curriculum level corresponding to the subterrain in which its starting node lies.
During training, if a system successfully traverses more than half of its assigned path, it is promoted to a higher curriculum level in the next episode. 
Conversely, if it covers less than a quarter of the path, it is demoted by one level. 
To mitigate catastrophic forgetting, systems that reach the highest level are reassigned to a random curriculum level.
At the start of training, all systems are initialized on low-difficulty paths, allowing them to first learn how to track the \textit{high-level} command in free space before progressing to more complex configurations.

It is worth emphasizing that the path planner has no knowledge of the full system’s length. 
Consequently, the sampled paths are guaranteed to be feasible for a single agent, but not necessarily for the entire system. 
This issue can arise at higher curriculum levels if a corner is too tight to navigate. 
Nevertheless, this does not impede the learning process, as reaching the highest curriculum level is not mandatory. 
Therefore, our method can be applied to systems with varying bar lengths without requiring prior knowledge of the full system dimensions.

\subsection{Action and Observation Spaces}
\label{sec:observations}
The \textit{high-level} actions $^{hl}_{obj}\textbf{a}$ correspond to the desired velocities of the agents' bases expressed in \textit{object} reference frame: $_{obj}(v_{x}, v_{y}, \omega_{z})_{base_i}$, $i = 1,2$.

The proprioceptive terms in the \textit{high-level} observations $^{hl}_{obj}\textbf{o}$ include: the object linear and angular velocities $_{obj}(v_{x}, v_{y}, \omega_{z})_{obj}$, the commanded velocity direction $^{hl}_{obj}\mathbf{c}$, the last applied action $^{hl}_{obj}\mathbf{a}^{t-1}$, the agents' base velocities $_{obj}(v_{x},\, v_{y},\, \omega_{z})_{base_i}$ and the relative yaw angle between object and agents' base frames: $_{obj}\psi_{obj-base_i}$.
The exteroceptive observations are represented by the height map of size $H\times W$ with resolution $h_{res}$ centered on and oriented as the \textit{object} frame. 
The heights of the points in this grid are collected in a vector of size $\mathbf{z}^t \in \mathbb{R}^{H\times W}$, which is concatenated with the proprioceptive observations.

\subsection{Policy Architecture}
\label{sec:policy}
As introduced in Sec. \ref{sec:prob_statement}, our architecture includes two nested layers.
The \textit{low-level} comprises two identical locomotion policies that track a planar SE(2) base velocity command $^{ll_i}_{base_i}\textbf{c}$ expressed in the corresponding agent base frame: $_{base_i}(v_x, v_y, \omega_z)_{base_i}$, $i = 1,2$.
Each policy outputs gait phase offsets and residual joint position targets, which are tracked by a joint-space PD controller that generates actuator torques.
These policies have already been trained and are kept frozen while training the \textit{high-level} controller.
Moreover, as shown in Fig. \ref{fig:architecture}, the \textit{low-level} commands $^{ll}_{obj}\textbf{c}$ correspond to the \textit{high-level} actions $^{hl}_{obj}\textbf{a}$.
The \textit{object-centric high-level} policy receives observations $^{hl}_{obj}\textbf{o}$ and commands $^{hl}_{obj}\textbf{c}$ relative to the whole system, leading to a fully centralized architecture.
This design choice is motivated by prior work \cite{quadruped_cent_dis_approaches, push_obstacle_RRT_planner}, where centralized architectures have demonstrated stronger performance than distributed or decentralized alternatives in similar settings.
\subsection{Rewards}
\label{sec:rewards}
In the following, unless otherwise specified, all quantities are expressed in the \textit{object} reference frame.
The parameters $w_i$ denote the weights of each term.
\subsubsection{Task-related Rewards}
\textbf{Tracking Command}: rewards the dot product between the commanded velocity direction and the normalized \textit{object} linear velocity along \textit{x} and \textit{y} axes:
\begin{equation}
    {R_t} = w_1(^{hl}\mathbf{c} \cdot ||{\mathbf{v}}_{{obj_{xy}}}||).
\end{equation}

\textbf{Align Object to Command}: symmetrically rewards the alignment of the \textit{object} \textit{y}-axis with the commanded velocity direction: 
\begin{align}
    {R_a} &= w_2\left(|\arctan(^{hl}\textbf{c}_y, ^{hl}\textbf{c}_x)| - \frac{\pi}{2}\right)^2, \hspace{20pt}
\end{align}

\subsubsection{Task-related Penalties}
\textbf{Distance from Obstacles}: penalizes short distances between the system's frames \textit{i: base1, base2, object} and their nearest obstacles:
\begin{align}
P_{dist_i} =
\begin{cases}
w_3 \cdot e^{\!\big(-\alpha (_W d_{min_i} - d_{s_i})\big)}, & \text{if} \hspace{3pt} _Wd_{min_i} < \delta, \\[2pt]
0, & \text{otherwise}, \\
\end{cases}
\end{align}
where $_Wd_{min_i} = \min(||_W\textbf{r}_{obs,i} - _W\mathbf{r}_{i}||)$, and $_W\textbf{r}_{i}$, $_W\textbf{r}_{obs,i}$ represent the positions of frame \textit{i} and its nearest obstacle.
Finally, $d_{s_i}$ and $\delta$ denote the safety radius and the distance threshold.

\textbf{Internal Forces}: penalizes the agents from trying to move in opposite directions along the \textit{object} \textit{y}-axis, minimizing the internal forces applied to the bar:
\begin{equation}
    P_{int forces} = w_4(e^{|^{hl}a_2 - ^{hl}a_5| - 1.0}).
\end{equation}

\textbf{Undesired Contacts}: 
penalizes the sum of net contact forces $F_{b_j}$ between specified system's bodies $b_j$ and the terrain, which includes the floor and the obstacles.
\begin{equation}
P_C = 
\begin{cases}
    w_5 \sum_{b_j} ||F_{b_j}||, &\text{if} \hspace{3pt} ||F_{b_j}|| > 1.0 \\[2pt]
    0, & \text{otherwise},
\end{cases}
\end{equation}

\textbf{Stand in Place}: penalizes the system when the cumulative displacements of the \textit{object} frame along both the \textit{x} and \textit{y} axes remain below a threshold $\tau$ over the previous $T_{stand}$ simulation steps.
\begin{equation}
P_{\text{stand}} =
\begin{cases}
w_6 \cdot e^{\!\Big(-\beta \sqrt{ \Delta_x^2 + \Delta_y^2 } \Big)} , 
& \text{if } \Delta_x < \tau \wedge \Delta_y < \tau, \\[2pt]
0, & \text{otherwise},
\end{cases}
\end{equation}
where $\Delta_x$ and $\Delta_y$ denote the $l_2$-norms of consecutive position increments along the 
\textit{x} and \textit{y} axes, respectively, accumulated over the history buffer.
\subsubsection{Regularization Penalties}
The regularization terms are meant to penalize the \textit{object} acceleration, $P_{obj \hspace{1pt} acc} =   w_7||\dot{\mathbf{v}}_{obj}||^2$, the action rate, $P_{a \hspace{1pt} rate} = w_8||^{hl}\mathbf{a}^{t-1} - ^{hl}\mathbf{a}^t||$, and the angular velocities of the agents bases' around their vertical axis, $P_{ang \hspace{1pt} vel} = w_9(\omega_{{base_1}_z}^2 + \omega_{{base_2}_z}^2)$

The final reward is computed as the sum of all the above terms.

\subsection{Terminations}
\label{sec:terminations}
We terminate the episode if any of the agents' bases are tilted over a certain angle $\gamma$ with respect to the world gravity vector, and if they are lower than a certain height threshold $T_h$.
These terms replace traditional terminations used in legged locomotion for undesired base contacts, which would negatively affect our collision-avoidance learning task, leading to an overly-conservative policy.
Finally, termination occurs once the maximum episode length is exceeded or the path goal node is reached.
These last two terms are outside the scope of the learning process and do not influence it negatively.

\section{Simulation experiments and results}
\label{sec:results}

\subsection{Setup}
\label{sec:sim}
The training of the proposed \textit{high-level} policy is conducted in the Isaac Lab simulation environment \cite{orbit}, using the Proximal Policy Optimization (PPO) algorithm \cite{ppo}.
Both actor and critic networks are implemented as multilayer perceptrons with hidden layers of size [128, 128, 128], and ELU as the activation function. 
Additionally, the actor network is provided with a final scaled \textit{tanh} layer, which constrains the actions to be in the range [$v_{min}$, $v_{max}$].
Finally, the system is trained with an object of length $L = 2$ m.
\subsection{Baselines comparisons}
\label{sec:mappo-opt-based}
In this section, we compare the proposed fully centralized policy with (i) a decentralized reinforcement learning baseline and (ii) an optimization-based method.
\subsubsection{Decentralized RL baseline}
We implement a decentralized Multi-Agent PPO (MAPPO)~\cite{mappo} algorithm following the Centralized Training with Decentralized Execution (CTDE) paradigm. 
During training, the critic has access to the global observations (Sec.~\ref{sec:observations}), while each agent receives only local observations. 
Specifically, each agent observes: the object linear and angular velocities, the desired object velocity, its own linear and angular velocities, its orientation relative to the bar, and its corresponding half of the system height map.
At execution time, the decentralized policy receives the object command defined in Sec.~\ref{sec:commands} and outputs desired base velocities expressed in the object frame as described in Sec.~\ref{sec:observations}.
The same \textit{low-level} controller from Sec. \ref{sec:policy} is then deployed.

The policy is trained and executed at 20 Hz, consistent with the centralized formulation (see Sec.~\ref{sec:hard_setup} for details).

\subsubsection{Optimization-based baseline.}
As an optimization-based baseline we implement the global planner presented in \cite{javier} and adapt it to our robotic system. 
It outputs a sequence of feasible system configurations connecting start and goal locations. 
Each configuration includes the object pose and the relative orientations of the agents with respect to the bar.
Agent trajectories are obtained by linearly interpolating between consecutive configurations.
Although this does not provide a fully realistic evaluation, we enforce a bound on the distance between successive poses so that they remain sufficiently close, making straight-line interpolation a reasonable approximation of the underlying motion.
To ensure a fair comparison, the optimization-based planner is restricted to the same local map knowledge as our policy, i.e., the area covered by the height map. 
The next configuration is computed via a probabilistic roadmap (PRM) within the visible region. 
We vary the number of sampled configurations $n_{\mathrm{samples}}$ (100 and 1500), while fixing the number of neighbors and interpolation steps to 5.

\begin{figure*}[ht!]
        \centering
        \includegraphics[width=0.8\textwidth]{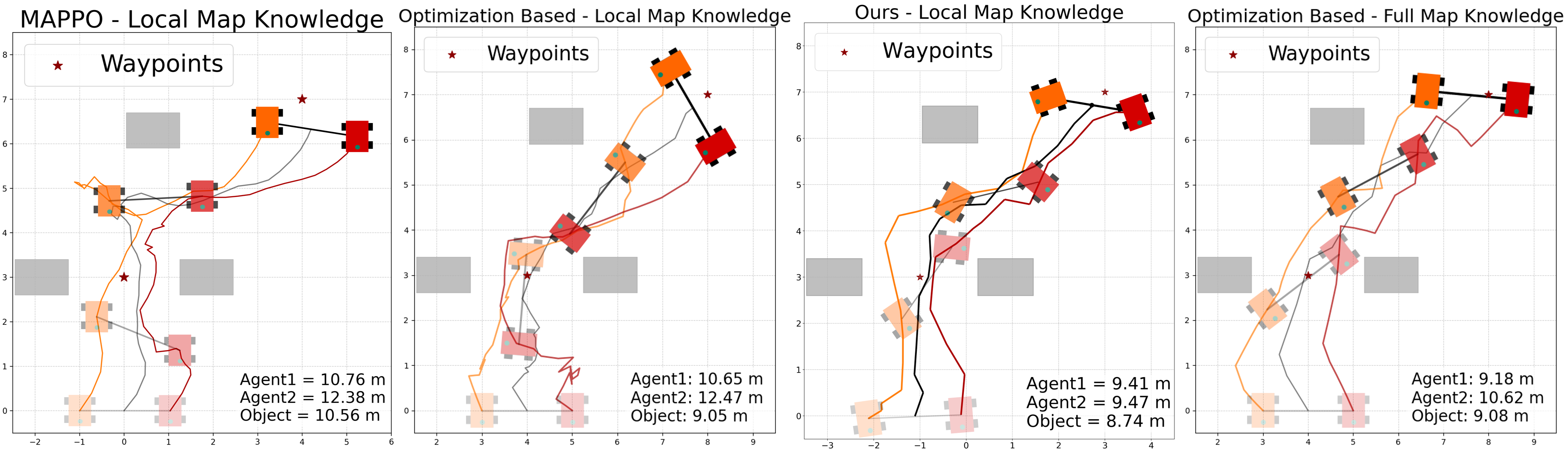}
        \caption{Trajectory comparison in the \textit{Boxes} scenario.
        From left to right: MAPPO, Optimization-based (local map, $n_{samples} = 1500$), Ours, and Optimization-based (full map, $n_{samples} = 1500$).
         The black line denotes the object path, colored lines represent the agents. 
         Red stars indicate intermediate waypoints.}
        \vspace{-17pt} 
        \label{fig:traj_comparison}
\end{figure*}

We evaluate all methods on three experiments:
\begin{itemize}
    \item \textbf{\textit{Boxes}}: Two obstacles are placed 3~m ahead of the initial configuration, forming a 2.5~m-wide corridor. A third obstacle, placed 2.5~m further, forms a second corridor. Two waypoints are defined: the first between the initial pair of obstacles, and the second 5.5~m away at a 45° angle.
    The setup is shown in Fig.~\ref{fig:traj_comparison}.
    \item \textbf{\textit{Empty}}: Identical waypoint configuration as above, but without obstacles.
    \item \textbf{\textit{Corridor}}: Two obstacles are placed 3~m ahead, forming a 2~m-wide corridor. 
    A single waypoint is positioned 7~m in front of the initial configuration.
\end{itemize}

Table~\ref{tab:avg_lengths_success_fullwidth} reports success rate (SR), execution frequency ($f$), and average path lengths for successful trials only. 
RL methods are evaluated over 100 trials of 70 s each, while the optimization-based baseline is evaluated over 10 trials with unconstrained duration, and includes a full-map variant ($n_{\mathrm{samples}}=1500$) for completeness. 
For the optimization-based method, increasing $n_{\mathrm{samples}}$ improves success but reduces frequency by one order of magnitude ($10^{-2}$→$10^{-3}$ Hz), making real-time deployment impractical. 
In the \textit{Empty} scenario, all methods achieve 100\% SR; however, only the RL policies maintain execution frequencies compatible with real-time operation.
In \textit{Corridor}, our method achieves the highest SR (99\%) among locally informed approaches, outperforming MAPPO (81\%) and the optimization-based planner (30\% / 90\%). 
In the more challenging \textit{Boxes} terrain, MAPPO degrades significantly (16\% SR), while the optimization-based planner reaches 80\% at the cost of reduced execution frequency and generally longer trajectories.

Our approach attains 99\% SR and the shortest average paths among locally informed methods in cluttered environments, demonstrating superior robustness and trajectory efficiency.
Moreover, compared to the optimization-based method with full map knowledge, our policy achieves near-identical success rates and comparable trajectory lengths across scenarios, despite relying solely on local perceptual information, while maintaining real-time execution at 20 Hz.
Finally, the consistent performance gains over the decentralized MAPPO baseline support our choice of a centralized architecture over a decentralized one.

An example of the trajectories obtained with the four compared methods for the \textit{Boxes} experiment is presented in Fig. \ref{fig:traj_comparison}.

\begin{table*}[!t]
\centering
\caption{Success rate (SR), execution frequency ($f$), and average path lengths for succesful trials of \textit{agent1}, \textit{agent2}, and the \textit{object} (mean $\pm$ std, in m) across experiments. 
The optimization-based planner is evaluated with local map knowledge for different $n_{\mathrm{samples}}$ and with full map knowledge using $n_{samples} = 1500$.}
\vspace{-2pt}
\label{tab:avg_lengths_success_fullwidth}
\begin{tabular}{llcccc|c}
\toprule
\textbf{} & \textbf{} &
\textbf{MAPPO} & \textbf{   \makecell{Opt.-based Local\\ $n_{samples}$ 100}   } & 
\textbf{   \makecell{Opt.-based Local \\ $n_{samples}$ 1500}   } & \textbf{Ours} & 
\textbf{   \makecell{Opt.-based Full \\ $n_{samples}$ 1500}   } \\
\midrule

\multirow{5}{*}{Empty}
& SR            & \textbf{100\%}   & \textbf{100 \%} & \textbf{100\%}   & \textbf{100\%}   & 100 \% \\
& $f$             & \textbf{20 Hz}               & $10^{-2}$ Hz & $10^{-3}$ Hz              & \textbf{20 Hz}              & -- \\
& $L_{agent_1}$ & 9.09 $\pm$ 0.11 & 10.10 $\pm$ 1.05 & 9.64 $\pm$ 0.74 & 8.00 $\pm$ 0.14 & 8.65 $\pm$ 0.01 \\
& $L_{agent_2}$ & 9.04 $\pm$ 0.08 & 9.67 $\pm$ 0.76 & 9.82 $\pm$ 0.81 & 8.48 $\pm$ 0.14 &  8.65 $\pm$ 0.01\\
& $L_{obj}$     & 9.02 $\pm$ 0.06 & 8.75 $\pm$ 0.26 & 8.98 $\pm$ 0.39 & 8.15 $\pm$ 0.05 &  8.65 $\pm$ 0.01 \\

\midrule

\multirow{5}{*}{Corridor}
& SR            & 81\%            & 30 \% & 90\%            & \textbf{99\%}    & 100 \% \\
& $f$             & \textbf{20 Hz }             & $10^{-2}$ Hz & $10^{-3}$ Hz             & \textbf{20 Hz}              & -- \\
& $L_{agent_1}$ & 7.49 $\pm$ 0.60 & 9.17 $\pm$ 2.08 & 8.50 $\pm$ 0.93 & 8.77 $\pm$ 0.30 & 9.06 $\pm$ 0.89 \\
& $L_{agent_2}$ & 8.04 $\pm$ 0.62 & 9.31 $\pm$ 1.24 & 8.21 $\pm$ 0.96 & 8.80 $\pm$ 0.50 & 9.38 $\pm$ 0.92 \\
& $L_{obj}$     & 7.07 $\pm$ 0.50 & 7.43 $\pm$ 0.65 & 7.26 $\pm$ 0.21 & 7.95 $\pm$ 0.21 & 7.63 $\pm$ 0.29 \\

\midrule

\multirow{5}{*}{Boxes}
& SR            & 16\%             & 30 \% & 80\%             & \textbf{99\%}     & 100 \% \\
& $f$             & \textbf{20 Hz}   & $10^{-2}$ Hz & $10^{-3}$ Hz              & \textbf{20 Hz}    & -- \\
& $L_{agent_1}$ & 15.42 $\pm$ 3.18 & 10.89 $\pm$ 1.03 & 11.27 $\pm$ 1.61 & \textbf{10.42 $\pm$ 1.87} & 12.20 $\pm$ 1.97 \\
& $L_{agent_2}$ & 16.62 $\pm$ 2.53 & 12.27 $\pm$ 1.79 & 12.01 $\pm$ 1.82 & \textbf{8.82  $\pm$ 1.69} & 12.47 $\pm$ 1.39 \\
& $L_{obj}$     & 14.06 $\pm$ 2.67 & 10.48 $\pm$ 1.26 & 9.62  $\pm$ 1.20 & \textbf{8.78  $\pm$ 1.54} & 11.03 $\pm$ 1.37 \\
\bottomrule
\end{tabular}
\end{table*}

\vspace{-5pt}
\section {Hardware experiments and results}

\subsection{Setup}
\label{sec:hard_setup}
Differently from simulation, where all system components run on the same machine, on hardware each locomotion policy executes on the robot’s on-board laptop, while the \textit{high-level} policy runs on a third external PC. 
To reproduce the simulation setup on hardware, we need to replicate: 1) the communication between the \textit{low-} and \textit{high-level} policies, 2) a shared fixed reference frame \textit{world}, and 3) a single elevation map centered on and aligned with the \textit{object} frame.

To enable communication between all the laptops, we use the ROS package \textit{nimbro network} \cite{nimbro}, which allows topic sharing via WiFi.
Moreover, we create a mesh network by deploying three \textit{Rajant} modules \cite{rajant}, one per agent and one connected to the external PC. 

To obtain a shared fixed reference frame \textit{world}, we align the \textit{x}-axes of \textit{base1} and \textit{base2} so that they point in the same direction (see Fig. \ref{fig:system}).
On \textit{agent1}, we then publish a new frame \textit{world} from \textit{map1} with an identity transform.
On \textit{agent2}, we publish \textit{world} from \textit{map2} with a translation of $-L$ along the \textit{y}-axis.
This procedure ensures that the two published \textit{world} frames overlap in the full-system localization environment, and the choice of agent used for alignment does not affect the outcome.
To define the reference frame \textit{object}, we identify its \textit{y}-axis as the normalized vector pointing from \textit{agent1} to \textit{agent2} position in \textit{world} frame.
The z-axis remains parallel to that of \textit{agent1} as, by design, the simulated system lacks a pitch joint between \textit{agent1} and the object.
This is required to close the \textit{sim2real} gap.
The \textit{x}-axis follows from the right-hand rule, and the frame is placed in the geometric center of the bar. 
Position, linear \textit{xy} and angular \textit{z} velocities of the object frame are calculated as follows, where all the quantities are expressed in \textit{world} frame:
\begin{align}
    \mathbf{r}_{obj} &= \frac{ \mathbf{r}_{base_1} + \mathbf{r}_{base_2}}{2} \\
    {\mathbf{v}}_{obj_{xy}} &= \frac{{\mathbf{v}}_{{base_1}_{xy}}  + {\mathbf{v}}_{{base_2}_{xy}}}{2} \\
    \omega_{obj_z} &=   \frac{\left({\mathbf{v}}_{{base_2}_{xy}} -{\mathbf{v}}_{{base_1}_{xy}} \right) \cdot \left(\frac{-{r}_{obj_y}, {r}_{obj_x}}{\|\mathbf{r}_{obj_{xy}}\|}\right)}{\|\mathbf{r}_{{obj}_{xy}}\|}  
\end{align}

Each agent builds its elevation map of size 8x8$m^2$ and 4 cm resolution using a Jetson Orin \cite{cupy}.
Since the obstacles are higher than the robots' bases, the agents cannot perceive the boxes' top surface, which results in empty cells in the elevation maps.
To circumvent this issue, we implement a \textit{max filter} on the elevation map, which fills the missing data with the highest value read among its neighbouring cells. \\
We decide to merge the maps collected by the two agents rather than using only one of them to avoid 1) perceiving the other robot as an obstacle, 2) having visual obstructions given by the other agent's position. 
Therefore, for each robot, we identify the portion of the map to be sampled and orient it as the object reference frame.
This submap is then down-sampled with a resolution of 30 cm, and the 3D coordinates of the points are published over a topic sent through \textit{nimbro} to the \textit{high-level} PC.

The communication of the height map through \textit{Rajants} from each agent to the \textit{high-level} PC takes on average 25 ms. 
This forces us to run the \textit{high-level} policy at 20 Hz, yielding an additional 50 ms.
Finally, the external PC sends the \textit{high-level} actions back to the agents via the \textit{rajants}, accumulating 25 ms more.
We define a safety distance from the nearest obstacle $d_{min}$ of 10 cm, which, together with the average delay $t_{delay}$ of 100 ms, limits each agent’s maximum linear velocity to $d_{min}/t_{delay} = 1$ m/s.
To account for occasional higher delays from the \textit{Rajant} modules, the \textit{high-level} actions are constrained within the range [-0.8, 0.8] using the final scaled \textit{tanh} layer described in Sec. \ref{sec:policy}, effectively limiting the maximum linear and angular velocities to $\pm 0.8$ m/s and rad/s, respectively.
\vspace{-5pt}
\subsection{Experiments}
\label{sec:hard_res}
\subsubsection{Waypoints-following}
\label{sec:waypoints_hardware_exp}
\begin{figure*}[ht]
    \centering
    \includegraphics[width=0.75\textwidth]{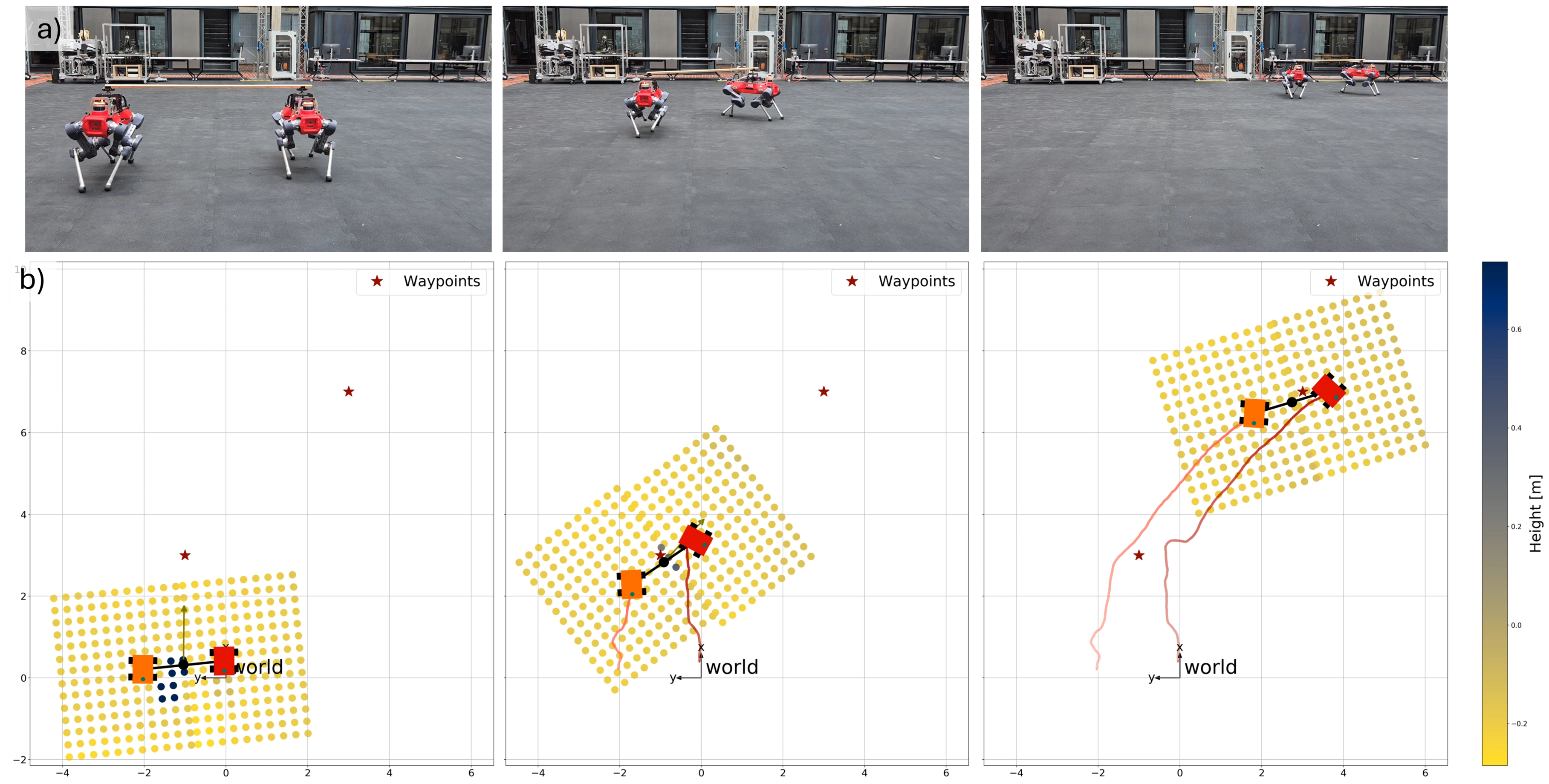}
    \caption{Waypoints-following on empty terrain experiment.
    Top: system configuration at the start and after reaching two predefined waypoints (red stars).
    Bottom: real-time exteroceptive observations, overlaid with the trajectories of the agents' bases and the position of the \textit{world} frame.}
    \vspace{-8pt}
    \label{fig:empty_wp}
\end{figure*}
\begin{figure*}[ht]
    \centering
    \includegraphics[width=0.75\textwidth]{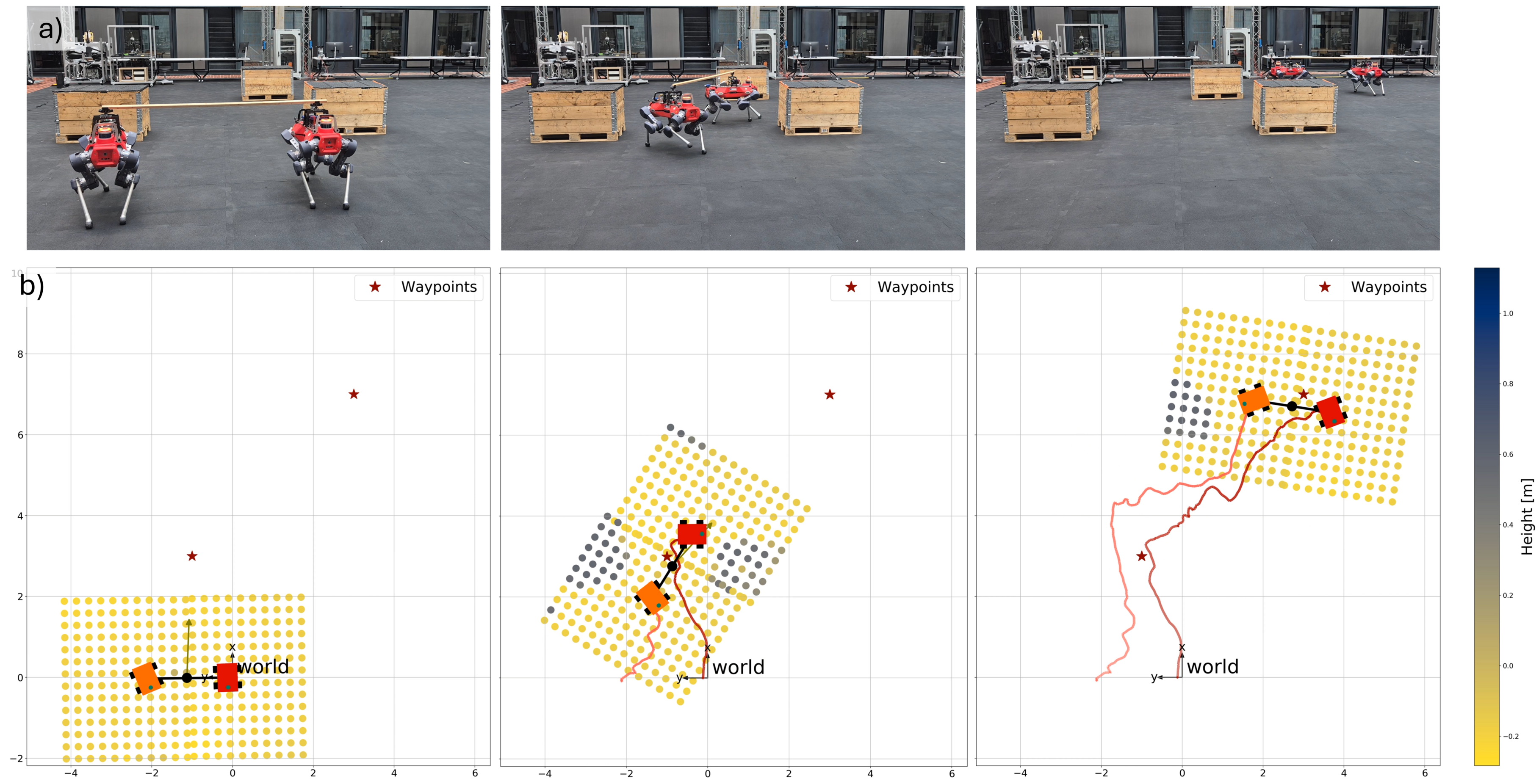}
    \caption{Waypoint-following on cluttered terrain experiment.
    Top: system configuration at the start and after reaching two predefined waypoints  (red stars).
    Bottom: real-time exteroceptive observations highlighting obstacle locations, overlaid with the trajectories of the agents' bases and the position of the \textit{world} frame.}
    \vspace{-16pt}
    \label{fig:boxes_wp}
\end{figure*}
We evaluate the differences in system behavior between free and cluttered environments replicating the experiments \textit{Boxes} and \textit{Empty} introduced in Sec. \ref{sec:mappo-opt-based}.

In all the following visualizations, the darker robot is referred to as \textit{agent1}, while the orange one as \textit{agent2}.
The behaviour and trajectories of the agents in empty space can be seen in Fig. \ref{fig:empty_wp}, while Fig. \ref{fig:boxes_wp} shows the case with obstacles.
Figures \ref{fig:empty_wp}b and \ref{fig:boxes_wp}b show the real-time exteroceptive observations collected by both agents and the trajectories computed by the agents' bases.
At times, due to base inclination, the bar enters the field of view of the depth cameras and is erroneously perceived as an obstacle. 
To account for this artifact during training, the height maps are augmented with noise to simulate such perception errors.
The right-most plots in these figures highlight clear differences between corresponding trajectories in the two experiments. 
This demonstrates that our policy adapts its behavior to the surroundings and that our system can locomote in unknown environments without requiring any \textit{a priori} map.
Furthermore, assuming the waypoint-defined trajectory of the \textit{object} frame is feasible, no path planner or prior knowledge of the length of the carried object is required.

In the subsequent experiments, the commanded velocity direction previously provided by waypoints-following is replaced by joystick input.

\subsubsection{Dynamic obstacle}

\begin{figure}[ht!]
        \centering
        \includegraphics[width=0.90\linewidth]{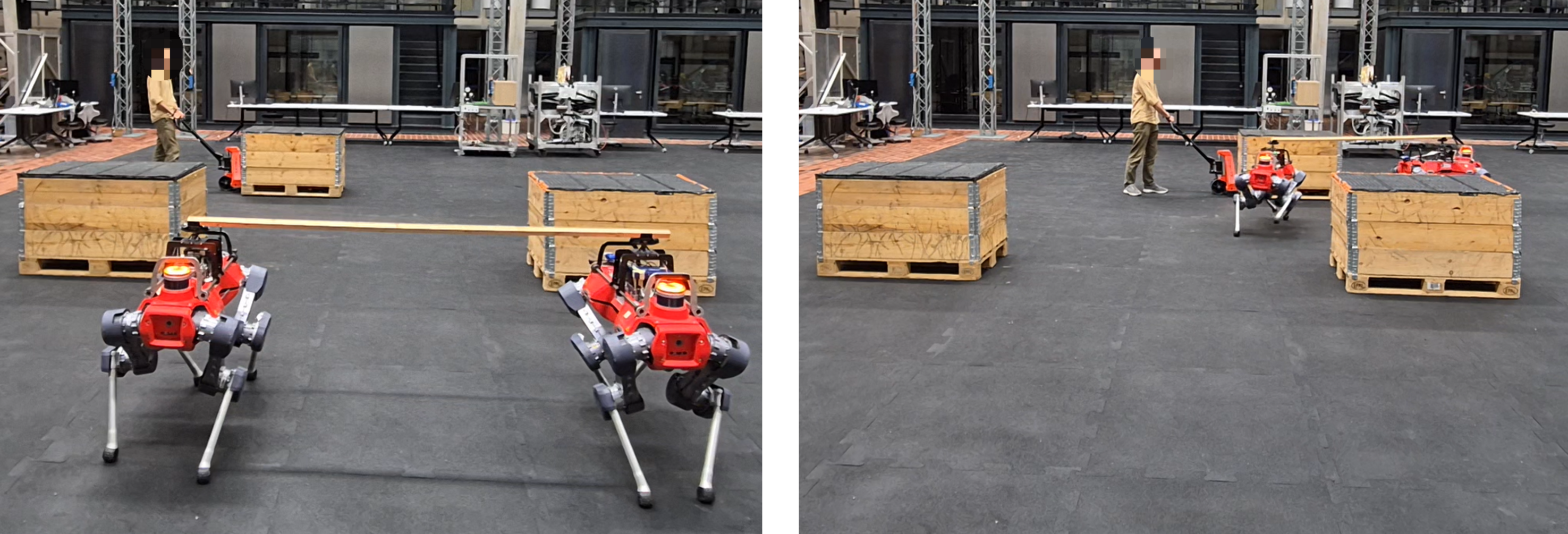}
        \caption{Placement of initial (left) and final (right) position of the box in the dynamic obstacle experiment.}
        \vspace{-10pt}
        \label{fig:dyno}
\end{figure}

Since our policy does not rely on prior map knowledge or precomputed paths, we assess its real-time reactivity by evaluating its response to a dynamic obstacle.
Fig. \ref{fig:dyno} illustrates the obstacle setup: two static boxes create a 2.5 m-wide corridor, while a third box, positioned 2.5 m further along, moves from an initial (left) to a final (right) location.
The system is commanded to traverse the corridor and then proceed diagonally to the right.
The detailed behaviour can be seen in the attached video, and closely resembles the one observed in the previous experiment.
The system successfully follows the commands without collisions, even when navigating near the dynamic obstacle.
The box is moved at a sufficiently low speed to allow updates of the elevation maps, which occur at 5 Hz \cite{cupy}.
\subsubsection{Corridors}
We further evaluate the policy by commanding the system to traverse corridors of decreasing width: 3.0 m, 2.5 m, and 2.0 m, the latter corresponding to the minimum width successfully crossed in simulation.
The corridor is defined by the two boxes in the first row of Fig. \ref{fig:boxes_wp}a, with the system commanded to move straight forward.
All trials are completed successfully, without any collisions, and the system exhibits consistent behaviour across all widths.

All the mentioned experiments and visualizations are available in the attached video.
\section{Conclusions}
We presented an RL-based policy for two coupled robots that is able to track a commanded velocity direction while also performing local collision-avoidance.
By benchmarking against optimization-based and decentralized RL methods, we demonstrate the comparative advantages of our approach.
Moreover, we validate our approach on hardware with two quadrupedal robots attached to a 2 m bar via spherical joints.
The system is tested in environments of increasing difficulty, from a corridor crossing to dynamic obstacle avoidance.
The system relies only on its onboard sensing and thus it does not require any pre-recorded map, as it is able to locomote in unknown environments. 
This also eliminates the need for a path planner that computes collision-free agents' trajectories, as this is taken care of by the policy.

Future work includes extending to more challenging terrains, such as slopes, stairs, and staircases with turns.
\section*{APPENDIX}
\label{sec:appendix}
\vspace{-12pt}
\begin{table}[H]
\centering
\renewcommand{\arraystretch}{1.2}
\setlength{\tabcolsep}{9pt}
\begin{tabular}{c|c||c|c}
\hline
$L$ & 2.0 m & $\alpha$ & 10.0 \\
$H\text{x}W$ & [4.0x6.0] m & $d_{s_{base_i}}$, $d_{s_{obj}}$ & 0.6 m, 0.2 m \\
$h_{res}$ & 0.3 m &  $\delta$ & 2.0 m \\
\cline{1-2}

$N_C$ & 50 &  $w_4$ & -0.2\\
$D_{max}$ & 0.1 & $w_5$ & -2.5\\  
$H_{obs} $ & 1.0 m & $w_6$ & -0.1\\  
$[s_{min}, s_{max}] $ & [1.0, 1.5] m & $\beta$ & 15.0\\  
\cline{1-2}

$n_p $ & 2000 & $T_{stand} $ & 10 \\  
$d_{s,wp} $ & 0.75 & $\tau$ & 0.15 m\\  
$N_P $ & 1500 & $w_7$ & -0.001  \\  
$[l_{min}, l_{max}] $ & [5.0, 12.0] m & $w_8$ & -0.0005\\  
$ R_{wp}$ & 0.5 m & $w_9$ & -0.1 \\  
\cline{1-4}

$w_1$ & -0.5 & $\gamma$ & 0.7 rad \\  
$w_2$ & -0.5 & $T_h$ & 0.15 m  \\  
\cline{3-4}
$w_3$ & -7.5 & $[v_{min}, v_{max}]$ & [-0.8, 0.8] \\ 
\cline{1-4}

\end{tabular}
\caption{Summary of all parameters introduced in Sec. \ref{sec:method}.}
\vspace{-12pt}
\label{tab:appendix_param}
\end{table}

\bibliographystyle{IEEEtran}
\bibliography{bibliography/references}

\end{document}